\documentclass[
	fontsize=11,
	headings=small,
	paper=a4,
	pagesize,
	DIV=calc,
]{scrartcl}

\usepackage{latexsym}
\usepackage{amssymb, amsfonts, amsmath, amsthm}
\usepackage{lmodern} 
\usepackage{microtype} 
 
\usepackage[scaled]{helvet} 
\usepackage{eulervm} 
\usepackage[T1]{fontenc}
\usepackage[utf8]{inputenc}
\usepackage{booktabs}
\usepackage{enumitem}
\usepackage{authblk}
\usepackage[margin=3cm]{geometry}
\usepackage{graphicx}
\usepackage{color}
\usepackage{xcolor}
\usepackage{caption}
\usepackage{tikz}
\usepackage{tcolorbox}
\usepackage[numbers, compress]{natbib}
\usepackage[
	pdfborder={0 0 0},
	pdftitle={Explaining Neural Networks with Reasons},
	pdfauthor={Levin Hornischer and Hannes Leitgeb},
	colorlinks=true,
	linkcolor=blue,
	citecolor=green,
	urlcolor=cyan,
]{hyperref}
\newcommand{\dox}{\mathrm{D}}
\newcommand{\RD}{\mathrm{RD}}

\title{\bfseries Explaining Neural Networks with Reasons}
\author[1]{\normalsize Levin Hornischer}
\author[1]{\normalsize Hannes Leitgeb}
\date{ }
\affil[1]{Munich Center for Mathematical Philosophy, LMU Munich}

\begin{document}

\maketitle
\renewcommand*{\thefootnote}{\fnsymbol{footnote}}

\begin{abstract}
\noindent
\textbf{Abstract}~~
We propose a new interpretability method for neural networks, which is based on a novel mathematico-philosophical theory of reasons. Our method computes a vector for each neuron, called its \emph{reasons vector}. We then can compute how strongly this reasons vector speaks for various \emph{propositions}, e.g., the proposition that the input image depicts digit 2 or that the input prompt has a negative sentiment. This yields an interpretation of neurons, and groups thereof, that combines a logical and a Bayesian perspective, and accounts for polysemanticity (i.e., that a single neuron can figure in multiple concepts).
We show, both theoretically and empirically, that this method is: 
(1)~grounded in a philosophically established notion of explanation, 
(2)~uniform, i.e., applies to the common neural network architectures and modalities, 
(3)~scalable, since computing reason vectors only involves forward-passes in the neural network, 
(4)~faithful, i.e., intervening on a neuron based on its reason vector leads to expected changes in model output, 
(5)~correct in that the model's reasons structure matches that of the data source,
(6)~trainable, i.e., neural networks can be trained to improve their reason strengths,
(7)~useful, i.e., it delivers on the needs for interpretability by increasing, e.g., robustness and fairness.\footnote{The source code will eventually be made available here: \url{https://github.com/LevinHornischer/ReasonsMethod}.} \medskip


\end{abstract}

\renewcommand*{\thefootnote}{\arabic{footnote}}
\setcounter{footnote}{0}

\section{Introduction}
\label{sec: introduction}

Neural networks, the drivers of the recent boom in artificial intelligence (AI), excel at learning patterns from data. However, they are also notoriously opaque: the parameters that they find during training are difficult to interpret in human-understandable terms. Solving this problem is the goal of AI interpretability research~\cite{DoshiVelez2017, Lipton2018, Murdoch2019, Molnar2025}. A prominent and rapidly growing approach is \emph{mechanistic interpretability}. It aims to analyze the internal mechanisms of the neural network in order to understand and improve it~\cite{Sharkey2025, Rai2025, Olah2023, Geiger2025}. At AAAI~2025, \citet{Chalmers2025} argued that this should be done specifically in terms of \emph{propositional attitudes}: using propositions, phrased in our language, that describe the AI system's goals and models of the world. Finding and logging these propositions is a research program that is ``highly nontrivial'' and ``we don't yet have any broad and reliable techniques''~\cite[p.~10]{Chalmers2025}.

In this paper, we suggest a first step. We build on a recent theory of reasons~\cite{Leitgeb2025} that formalizes the language of \emph{reasons}, which we ordinarily use to make sense of the world and the mechanisms in it---be it physical processes, the behavior of others, or engineering artifacts. Based on this theory, we develop a new interpretability method for neural networks. It makes sense of individual neurons, and groups thereof, as epistemic reasons that favor certain propositions with certain numerical strengths. We compute, for each neuron, its reasons vector, from which we can compute how much it speaks for each proposition. For example, for a neural network solving the MNIST task, we will compute, e.g., that this particular neuron speaks with strength 2.36 for the proposition that the input image depicts digit~3. Or in an LLM, we can compute that this group of five neurons speaks most strongly for the prompt having a positive sentiment. 

We find that our reasons method satisfies ten desiderata for interpretability that we identify in the literature (section~\ref{sec: background}). The method applies to both individual neurons and groups thereof, and it is rooted in a fundamental conceptual framework of making sense of the world (section~\ref{sec: methodology}). The method can account for polysemanticity, since a single reason can speak for multiple propositions. It connects to the logico-symbolic tradition of understanding cognition by associating neurons with propositions, and it also connects to the Bayesian tradition by describing how neurons update subjective probabilities. In experiments (section~\ref{sec: experiments}), we see that the method applies across the common neural network architectures and modalities. It is scalable, since computing the reasons vector only involves forward-passes. The method is faithful, i.e., intervening according to the reasons brings about the expected change in behavior; and it is correct in the sense that the reason structure of a well-trained model matches that of the world. Reasons not only interpret trained models, but we can also train a model via backpropagation to improve its reasons strengths, and this also increases robustness and fairness.

\section{Background: Desiderata for interpretability}
\label{sec: background}

We identify desiderata for any interpretability method that aims for mechanistic---or even propositional~\cite{Chalmers2025} ---interpretability. Afterward, we discuss them and the respective literature.

\begin{enumerate}
    \item \emph{Understandable}: 
    \label{itm: int req understandable}
    The interpretation should be in human-understandable terms. 

    \item \emph{Local and distributed}:
    \label{itm: int req local and distributed}
    The interpretability method should interpret a neural network in both a local (individual neurons) and a distributed (groups of neurons) way.

    \item \emph{Mechanism-compatible}: 
    \label{itm: int req mechanism compatible}
    The interpretation should reflect: (a)~the encoding of inputs, (b)~the real-valued activations of neurons across different inputs, (c)~the decoding of outputs, and (d)~how neurons interact via weights and activation functions with other neurons. 

    \item \emph{Uniform}: 
    \label{itm: int req uniform}
    The method should apply to all neural network architectures and data modalities.
    
    \item \emph{Scalable}: 
    \label{itm: int req scalable}
    The method should work both for small and large neural networks.

    \item \emph{Transparent}: 
    \label{itm: int req transparent}
    The interpretability method should not require further interpretation of its results or black-box training.

    \item \emph{Grounded}:
    \label{itm: int req grounded}
    The method should provide a philosophically deep notion of interpretation that supports comparisons between a neural network, its interpretation, and reality.
    
    \item \emph{Faithful}: 
    \label{itm: int req faithful}
    The method should assign interpretations that represent faithfully, i.e., which track features of the network under relevant interventions. 
    
    \item \emph{Correct}: 
    \label{itm: int req correct}
    The method should assign interpretations that represent correctly, i.e., where the structure of the interpretations tracks the intended structure of reality (the data).

    \item \emph{Useful}: 
    \label{itm: int req useful}
    The method should deliver on the needs for interpretability, i.e., trust, causality, transferability, fairness, privacy, robustness/reliability, recourse, and debugging.
    
\end{enumerate}

While~\ref{itm: int req understandable} is uncontested, the literature is undecided on~\ref{itm: int req local and distributed}: whether it is individual neurons that should be interpreted (as in the first neural networks~\cite{McCulloch1943}) or rather groups thereof (distributed representation). See, e.g., \cite{Smolensky1988, Olah2020, Garcez2023} for discussion.
Although interpretations of individual neurons have been suggested in some cases~\cite{Olah2017, Bau2020}, a \emph{complete} localist representation is difficult, since neurons often are polysemantic, i.e., participate in several concepts---they are in `superposition'~\cite{Smolensky1988, Olah2020, Elhage2022}. But~\ref{itm: int req local and distributed} asks for as much of a \emph{partial} localist representation as possible, which ideally explains how distributed representations are built up. 
Our reasons method provides this: the reasons vector is a local representation since it is associated with a specific neuron, and it also partial since it `pushes' into different conceptual dimensions rather than a single one (cf.\ superposition). Aggregating the reasons vectors of a group of neurons builds a distributed representation of the group.

Regarding~\ref{itm: int req mechanism compatible}, part~(b) precludes an interpretation of a neuron's activation as a classical truth-value, but it allows more complex interpretations in terms of truth and falsity~\cite{Herrmann2025, Marks2024}. The reason method will interpret activations as providing the components of the reasons vectors. Regarding~(d),~\cite{Wang2022} discusses desiderata for circuit discovery. For us, weights and activation functions determine the reasons vector of a neuron based on the reasons vectors of the neurons in the preceding layer.

Regarding~\ref{itm: int req uniform}--\ref{itm: int req transparent}, one of the most prominent approaches to mechanistic interpretability, \emph{sparse auto-encoders} (SAEs)~\cite{Cunningham2023, Bricken2023}, required much work to scale~\cite{Templeton2024}. Our reasons method only requires forward passes of the model, while SAEs require extensive training, which make them expensive to compute and difficult to evaluate~\cite{Gao2025, Heap2025}.

\begin{figure}
    \begin{minipage}{0.59\linewidth}
    \centering
    \begin{small}
    \begin{tikzpicture}
    \tikzset{boxstyle/.style={draw=blue, fill=blue!10, rounded corners}}
    
    \node[boxstyle] at (0,0) (A) {neural network};
    \node[boxstyle] at (5,0) (B) {reality/data};
    \node[boxstyle] at (2.5,1.5) (C) {interpretation};
    \draw[<->, >=latex] (A)--(B) node[midway, above] {\emph{accurate}};
    \draw[<->, >=latex] (A)--(C) node[midway, above left] {\emph{faithful}};
    \draw[<->, >=latex] (B)--(C) node[midway, above right] {\emph{correct}};    
    \end{tikzpicture}
    \end{small}
    \end{minipage}
    \begin{minipage}{0.4\linewidth}
    \begin{center}
    \begin{tabular}{c | c c c}
                & $u_1$     & $\ldots$  & $u_n$     \\ \hline
    $w_1$       & $a_{11}$  & $\ldots$  & $a_{1n}$  \\
    $\vdots$    & $\vdots$  &           & $\vdots$  \\
    $w_{2^m}$       & $a_{{2^m}1}$  & $\ldots$  & $a_{{2^m}n}$  \\
    \end{tabular}
    \end{center}
    \end{minipage}
    \caption{\emph{Left}: The triangle of interpretability. \emph{Right}: The activation matrix.} 
    \label{fig: triangle of interpretability and activation matrix}
\end{figure}
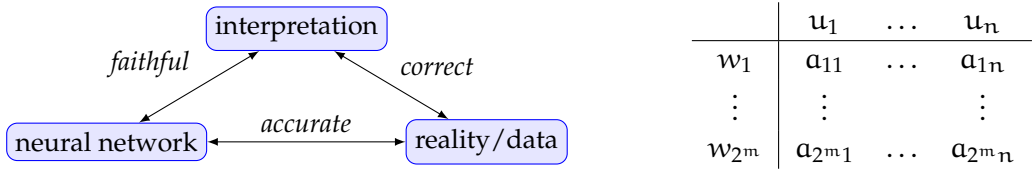

Desiderata~\ref{itm: int req grounded}--\ref{itm: int req correct} are explained in what we call the \emph{triangle of interpretability} in figure~\ref{fig: triangle of interpretability and activation matrix}~(left).
The interpretability method should connect three components: 
(1)~the neural network;
(2)~the human-understandable interpretations of the network parameters; 
(3)~the reality, which is available via data.
The connection (1)--(3) is measured as accuracy: it requires no interpretability since it just demands that the network's behavior matches reality.
The connection (1)--(2) requires that the network's internal \emph{mechanism} is captured by the interpretation. This is known as \emph{faithfulness} and tested in terms of interventions~\cite{Milliere2024b, Geiger2025, Harding2023, Vig2020, Zhang2024, Meng2022}. Intervening on the network parameters according to the interpretation should result in the expected change in behavior.
While accuracy requires \emph{behavioral correctness}, connection (2)--(3) requires \emph{mechanistic correctness}. Given an interpretation of the network, we not only want it faithfully representing the internal mechanisms, we also want that the model's mechanisms match those of reality. In section~\ref{ssec: interpreting a classic}, we operationalize this notion for our reasons interpretation via dimensionality reductions.

Finally,~\ref{itm: int req useful} is commonplace~\cite{Lipton2018, DoshiVelez2017}, and we will find that reasons improve robustness and fairness.

\section{Methodology: Reasons and the Reasons Method}
\label{sec: methodology}

We review the philosophical understanding of reasons and then the recently axiomatic theory of reasons~\cite{Leitgeb2025}. Afterward, we apply it to develop our reasons method for interpreting neural networks. 

\paragraph{Philosophy of reasons} Talk of reasons for action (practical reasons) and reasons for belief (epistemic reasons) is omnipresent in everyday communication and much researched in philosophy~\cite{Howard2024}. Given our focus on interpretability, we consider epistemic reasons. Different reasons may support the same proposition, and one and the same reason may support different propositions. Moreover, reasons may do so with different strengths. Having a reason for a proposition $A$ is a particular propositional attitude, which is required for an agent to believe $A$ and which increases the probability the agent assigns to $A$ (proportional to the reason's strength). Since reasons can act against each other and even defeat each other~\cite{Horty2012}, a rational agent needs to aggregate all available reasons before forming a belief on their basis. One understanding of reasons is as `epistemic forces' that should aggregate additively much like physical forces do. This suggests that reasons might have a vector structure, and aggregation and attenuation of reasons is vector addition and scalar multiplication, respectively.

\paragraph{Theory of reasons}
The theory of reasons~\cite{Leitgeb2025} formalizes these philosophical ideas. It axiomatizes the following primitive notions:
\begin{itemize}
\item `$x$ is a direct epistemic reason of strength $\alpha$ for proposition $A$' (written $R(x, A, \alpha)$).
E.g., $x$ might be a strong reason for \emph{there will be rain} presented by black clouds. 

\item `$x \circ y$ is the aggregation of reasons $x$ and $y$'.
E.g., $y$ could be another reason presented by a weather forecast, and $x \circ y$ would be the aggregation of the two reasons.

\item `$b * x$ is the result of updating the agent's current beliefs $b$ with (the available) reason $x$'.
E.g., if $x$ speaks strongly for $A=$~\emph{there will be rain}, the posterior subjective probability $b * x(A)$ should be significantly greater than the prior probability $b(A)$.
\end{itemize}
For a given reason $x$, it is not necessarily the case that for every proposition $A$ there is an $\alpha$, such that $R(x, A, \alpha)$.
To generalize $R$ to all propositions $A$, additional probabilistic weighing is required. For that purpose, the theory allows for the definition of a more inclusive reason relation:
\begin{itemize}
\item `$x$ is a doxastic reason of strength $\alpha$ for proposition $A$ that is inferred relative to belief $b$' (written $S(x, A, \alpha, b)$).
E.g., given my belief that tonight's party is sensitive to bad weather, black clouds speak with high strength for the party being canceled.
\end{itemize}
The axioms for these primitive notions include, e.g., the additivity of $\circ$ with respect to $R$: $R(x \circ y, A, \alpha)$ iff there are $B, C, \beta, \gamma$ such that $R(x,B,\beta)$, $R(y,C,\gamma)$, $A = B \cap C \neq \emptyset$, and $\alpha = \beta + \gamma$. In contrast, $\circ$ is not generally additive with respect to $S$.
The main mathematical result is that, surprisingly, the models of the overall axiomatic theory of reasons are unique up to a multiplicative constant $c > 0$. Hence we will work here directly with the models given for $c = 1$. 

The models of the reasons theory are determined by a choice of a finite set $W = \{ w_1, \ldots , w_{2^m} \}$, for some positive integer $m$. Its elements will be called \emph{possible worlds} or \emph{samples}. (They will be, as we will soon see, the situations in which the neural network can be applied, e.g., input-label pairs.) Together with the powerset $\mathcal{P}(W)$, it forms a measurable space. The elements of $\mathcal{P}(W)$ are called \emph{propositions} or \emph{events}---i.e., a proposition is identified with the set of possible worlds at which it is true. The \emph{negation} or \emph{complement} of $A$ is given by: $A^c := W \setminus A$.\footnote{The terms `possible worlds', `propositions', `negation' are used in philosophy, while `sample', `event', `complement' are used in statistics.} A \emph{belief} is a probability measure on $(W, \mathcal{P}(W))$. A \emph{reason} $x$ is a vector in $\mathbb{R}^{2^m}$. Intuitively, $x$ is the reason that speaks with strength $x_k$ for $w_k$ being the actual world---i.e., $R(x,\{w_k\},x_k)$. 
Given a proposition $A \subseteq W$, the \emph{elementary reason} for $A$, written $\mathrm{el}_A$, is the vector in $\mathbb{R}^{2^m}$ which is $1$ at component $k$ if $w_k \in A$ and $-1$ otherwise.
Reason aggregation $\circ$ is vector addition. (So aggregations of elementary reasons need not be elementary again.) 
Given a probability measure $b$ on $W$ and a reason $x \in \mathbb{R}^{2^m}$, the \emph{update} of $b$ by $x$ is the probability measure defined by (for $A \subseteq W$): 
\begin{equation}
\label{eqn: star update}
    b * x (A) 
    :=
    \frac{
        \sum_{k = 1}^{2^m} e^{x_k} b (A \cap \{ w_k \})
    }{
        \sum_{k = 0}^{2^m} e^{x_k} b (\{ w_k \})
    }.\footnote{If $b$ is the uniform measure, $b * x$ is the well-known softmax of $x$.}
\end{equation}
The \emph{doxastic reason strength} $\alpha$ with which $x$ speaks for a proposition $A$ relative to $b$ (i.e., $S(x, A, \alpha, b)$) is defined by:
\begin{equation}
\label{eqn: reasons strength}
    \alpha 
    := 
    \dox(x, A, b, W) 
    :=
    \frac{1}{2}
    \log\Big(
    \frac{
        b * x (A) / b * x (A^c)
    }{
        b (A) / b(A^c)
    }
    \Big).
\end{equation}
which is defined if, and only if, the proposition $A$ is \emph{nontrivial}, i.e., $0 < b(A) < 1$.

\paragraph{Reasons method for interpretability}
Our reasons methods uses the reasons theory to interpret neurons and groups thereof as follows. 
The main conceptual choice is the set $W  =\{ w_1, \ldots, w_{2^m} \}$ of situations in which the neural network can be applied. For example, in an image classification task, this could be input-label pairs. (We will see many more examples in section~\ref{sec: experiments}.) Given a neuron $u$ of the neural network, its \emph{reasons vector} $r_u \in \mathbb{R}^{2^m}$ has, as value at component $k$, the activation that neuron $u$ has in the possible world $w_k$.\footnote{We do not use the variable `$x$' since it commonly refers to the input to the neural network.} In the example, if $w = (x,y)$ is an input-label pair, then $r_u$'s value at $w$ is simply the activation of neuron $u$ after inputting $x$ to the neural network.

Once the reasons vector $r_u$ is computed, we can use it to interpret the neuron $u$ in two ways---in line with logical and Bayesian tradition, respectively.  
(1)~\emph{Logico-symbolically}: The neuron $u$ represents the proposition $A$ consisting of those possible worlds at which $r_u$ has a positive value.  
(2)~\emph{Probabilistically}: Relative to a prior probability measure $b$ on $W$, the neuron $u$ represents the probability distribution $b * r_u$.
We get a combination of both---which we hence will use below---via the reasons theory.
(3)~\emph{Strength-based}: Relative to a prior probability measure $b$ on $W$, the neuron $u$ represents a strength profile, i.e., how much it speaks for and against any nontrivial proposition. 
For example, if $l$ is a label in the classification task, the proposition `The input has label $l$' is the set $A = \{ (x,y) \in W : y = l \}$, and neuron $u$ speaks for it with strength $\dox(r_u,A,b,W)$.

Finally, we interpret a group of neurons $u_1, \ldots, , u_n$ by the aggregated reasons vector $r := r_{u_1} + \ldots + r_{u_n}$. This choice is corroborated by the general result of the reasons theory that update and aggregation commute: $b * r$ is the same as the convex combination of the $b * r_k$'s.

\section{Experiments}
\label{sec: experiments}

To experimentally test our new interpretability method, we apply it in a wide range of tasks: involving different architectures (convolutional neural networks, multi-layer perceptrons, and transformer-based LLMs) and different modalities (images, tabular data, and text).

\subsection{Interpreting a classic: LeNet for MNIST}
\label{ssec: interpreting a classic}

Given the method's novelty, we first test it on a task---the MNIST task---with the classic architecture that solved it: the convolutional neural network \emph{LeNet} \cite{LeCun1995}.
The task is to classify images of handwritten digits according to which digit ($0, \ldots, 9$) they depict. We train the LeNet architecture on the MNIST training set and achieve $> 99\%$ accuracy on the test set (details in appendix~\ref{app: exp 1}).

\begin{figure}
    \centering
    \includegraphics[width=0.49\linewidth]{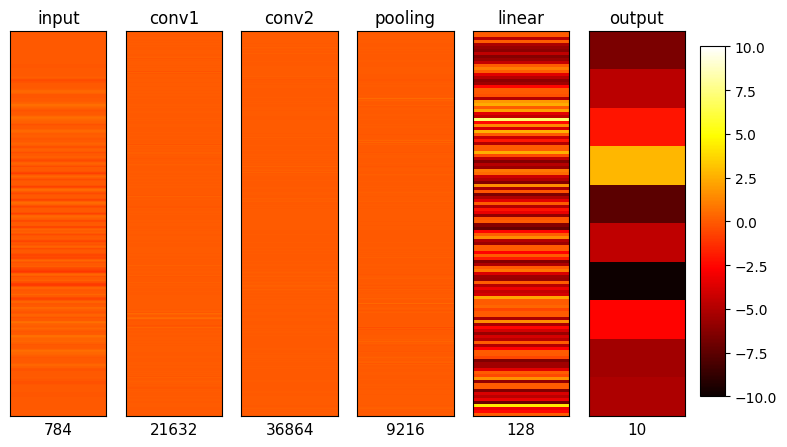}
    \includegraphics[width=0.49\linewidth]{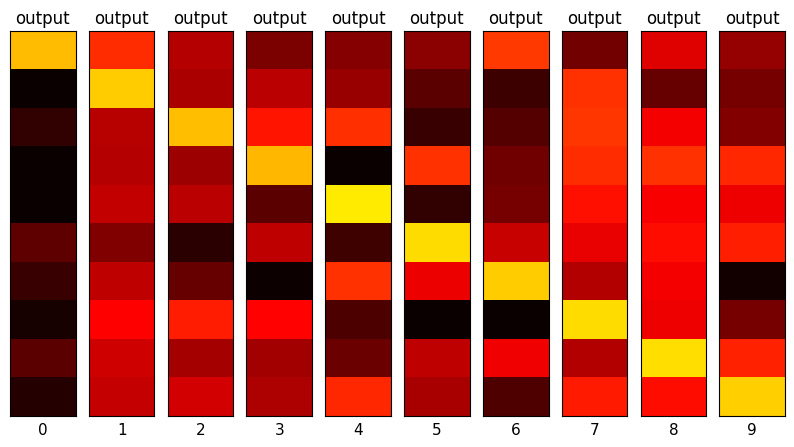}
    \caption{\emph{Left}: For each neuron in the different layers of LeNet, the strength with which it speaks for (positive) or against (negative) the proposition `The input depicts digit 3'. The number below the bars indicates the number of neurons in the layer. 
    \emph{Right}: For each digit $d$ (shown below each bar), the reasons strength of the output neurons speak for `The input depicts digit $d$'.}
    \label{fig: reason strengths}
\end{figure}

\paragraph{Reason strengths}
To apply our reasons method, we choose the possible worlds as input-label pairs.\footnote{We could add more information to a world: e.g., who wrote the digit; when and where it was written; or an intended label in addition to the correct label (in case, say, someone wrote what looks like a `7' but meant a `1').} We sample a set $W$ of $1024$ such pairs $(x,l)$ from the test set, so the model has not seen them.\footnote{Appendix~\ref{app: exp 1} establishes statistical robustness and shows no qualitative difference to using the training set.} The propositions of interest are `The image depicts digit $d$', i.e., $A_d := \{ (x,l) \in W : l = d\}$. 

Now, for each neuron $u$ in the trained LeNet model, we can compute its reasons vector $r_u := ( u(x) : (x,l) \in W)$, where $u(x)$ is the activation of neuron $u$ in the model on input $x$. By taking the uniform measure $b$ on $W$, we can compute, for each proposition $A_d$, the strength $\dox(r_u,A_d,b,W) $ with which neuron $u$ speaks for the proposition $A_d$. This is shown, for $d = 3$, in figure~\ref{fig: reason strengths}~(left). 
We observe fairly low reason strengths in earlier layers and stronger ones (either positive or negative) in the later layers. This is in line with CNNs possessing a hierarchy of features: with earlier layers corresponding to low-level features such as basic shapes, while later layers correspond to more abstract features~\cite{Zeiler2014}.
Focusing on the output neurons, figure~\ref{fig: reason strengths}~(right) shows their reasons strength for the different digits. As desired, for each digit $d$, the output neuron corresponding to $d$ strongly speaks for the proposition `The image depicts digit $d$', while the other output neurons strongly speak against it.

When it comes to interpreting groups of neurons, the layers make a natural choice. In appendix~\ref{app: exp 1}, figure~\ref{fig: app layerwise update} shows how the layers (after aggregating the reasons vectors of their neurons) update an initially uniform prior probability distribution over the possible worlds. Even though we start with the uniform measure and use a balanced dataset, the input layer introduces some bias among the worlds---and this bias is amplified by later layers.

\paragraph{Faithfulness}
Next, we test the faithfulness of our interpretation via \emph{causal interventions}~\cite{Harding2023, Geiger2025, Milliere2024b} or, more precisely, \emph{activation patching}~\cite{Vig2020, Zhang2024, Meng2022, Geiger2021}. Specifically, we test this in two versions, for the hidden linear layer of our trained LeNet model.

\begin{figure}
    \centering
    \includegraphics[width=0.49\linewidth]{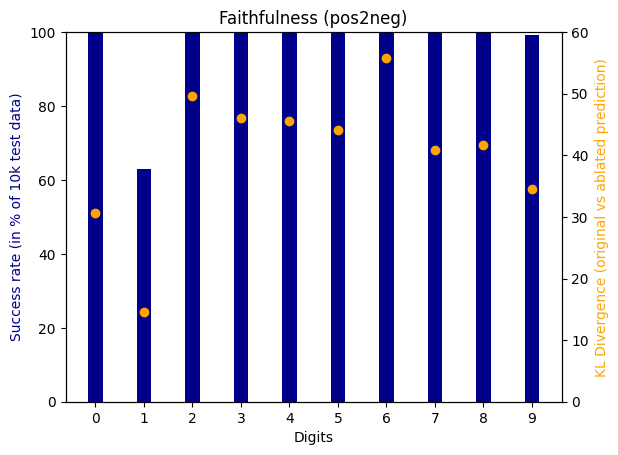}
    \includegraphics[width=0.49\linewidth]{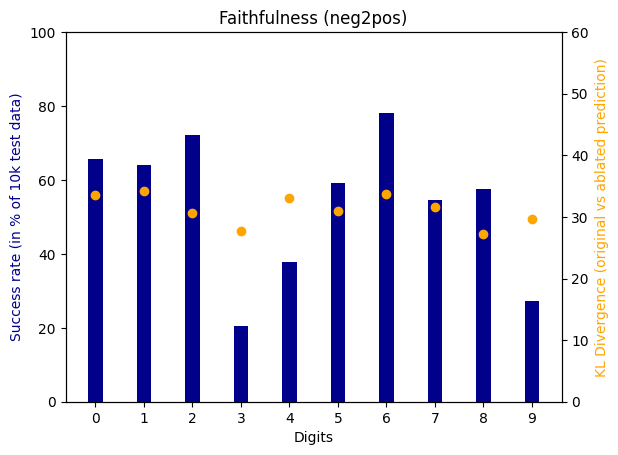}
    \caption{\emph{Left}: intervening on neurons speaking against a digit to flip the prediction away from that digit. \emph{Right}: intervening on neurons speaking for a digit to flip the prediction to that digit.}
    \label{fig: faithful}
\end{figure}

\emph{Version 1: pos2neg}. Fixing a digit $d$, we go through the test dataset considering images $x$ that are labeled with $d$. We input $x$ into the model, which, due to its high accuracy, will classify $x$ almost always correctly as $d$. Now we consider the activations of the 20 neurons in the linear layer that most strongly speak against digit~$d$. We intervene and set their activations to $a' := m - 3 a$, where $m$ is that neuron's mean activation and $a$ is its current activation.\footnote{Taking the mean---aka \emph{mean ablating}---effectively `knocks out' the neuron; and it does so better than \emph{zero ablating}, i.e., setting the neuron to zero~\cite{Wang2022, Zhang2024}. Adding $-3a$ points the neuron in the opposite direction.} From these intervened activations in the linear layer, we forward propagate to calculate the intervened model output. It is a success if the model now predicts a digit different from $d$. 
Figure~\ref{fig: faithful}~(left) shows that, for all digits except $1$, we have a $100\%$ success rate. It also shows (as orange dots) the KL divergence between the originally outputted probability distribution over the digits and the one after intervention.

\emph{Version 2: neg2pos}. Fixing a digit $d$, we now consider test images $x$ that are \emph{not} labeled with $d$. We input $x$ into the model and now consider the activations of the 20 neurons in the linear layer that most strongly speak \emph{for} digit~$d$. We intervene to set their activations to $a' := m - 5 a$ and calculate the intervened model output. It is a success if the model now predicts digit $d$. 
Note that this is much harder: intervening to do \emph{anything} else is easier than doing something \emph{specific}. 
Still, figure~\ref{fig: faithful}~(right) shows that, for all digits except $3,4,9$, we have an approximately $60\%$ success rate and, for all digits, a KL divergence of around $30$. In the next subsection, when we train the model's reasons, we see that these success rates improve---thus further corroborating faithfulness.

\paragraph{Correctness} 
We need to operationalize the idea that the reasons structure of the neural network should match the reasons structure of the world. Inspired by theories of scientific representation~\cite{Frigg2021}, we measure how much the representational similarity between possible worlds matches their objective similarity.
This is done via the activation matrix in figure~\ref{fig: triangle of interpretability and activation matrix}~(right).
Given neurons $u_1, \ldots , u_n$ and possible worlds $w_1, \ldots, w_{2^m}$, the value $a_{ij}$ is the activation of neuron $u_j$ at world $w_i$. Thus, the $j$-th \emph{column} is the reasons vector of neuron $u_j$, and we call the $i$-th \emph{row} the \emph{reasons-character} of the world $w_i$ (cf.\ $C^*$ algebras).
Two worlds are \emph{internally similar} if their reasons-characters are close as vectors in $\mathbb{R}^n$: the reason structure of the neural network almost cannot tell these worlds apart. Two worlds are \emph{externally similar} if they have the same objective properties, i.e., the same label.
Correctness requires that internal similarity typically (i.e., \emph{defeasibly}) entails external similarity. Thus, worlds with the same label should form clusters in the space $\mathbb{R}^n$ of reasons-characters. 

\begin{figure}
    \centering
    \includegraphics[width=0.49\linewidth]{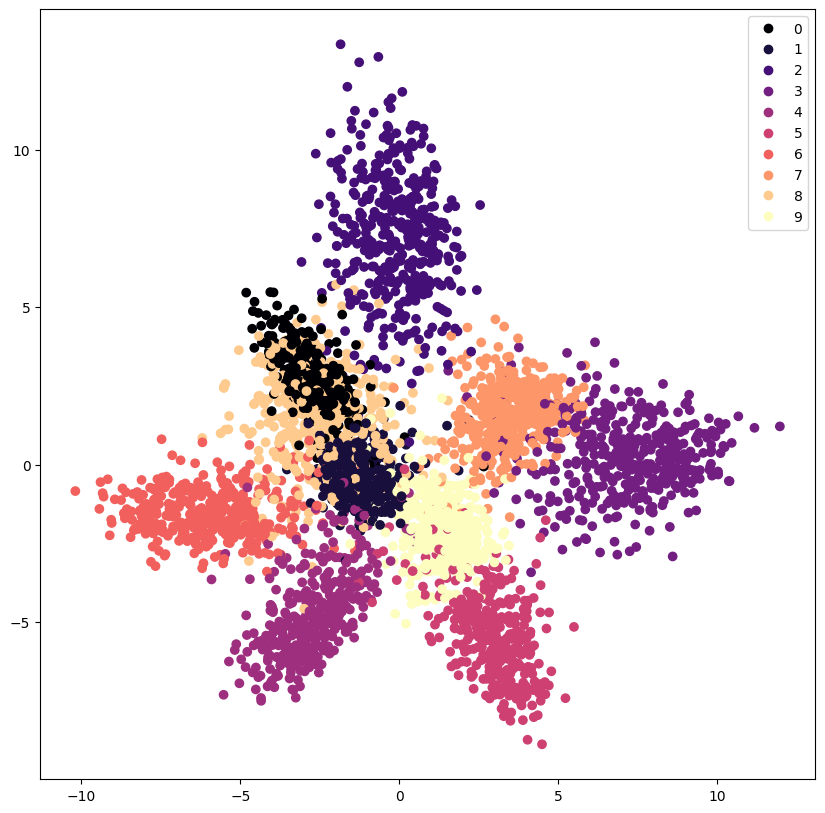}
    \includegraphics[width=0.49\linewidth]{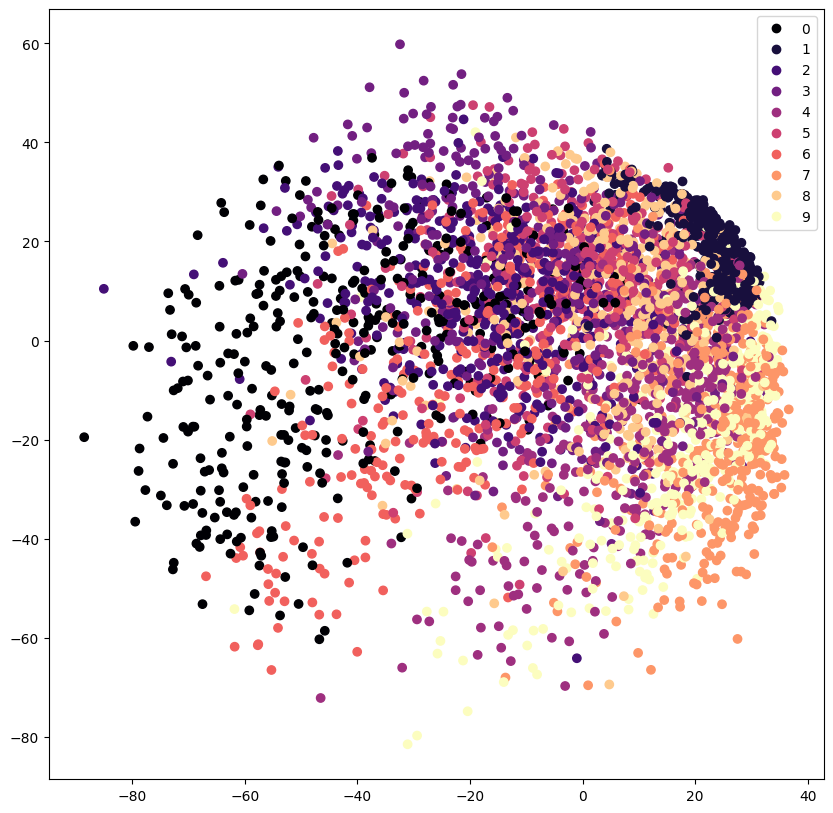}
    \caption{\emph{Left}: After clustering together worlds (using PCA) that are internally similar according to the neurons in the hidden linear layer, they also are externally similar, i.e., have the same label. \emph{Right}: This is not yet true for neurons in the first convolutional layer.}
    \label{fig: correctness}
\end{figure}

To observe potential clusters, we need to reduce the dimension from $n$ to $2$. So we perform a Principal Component Analysis (PCA). (Appendix~\ref{app: exp 1} shows similar results for t-SNE and UMAP.) We sample $2^{12}$ worlds from the test set and consider the neurons of the hidden linear layer. We associate each of the 10~labels with a color. So if correctness holds, we should find monochromatic clusters---which is the case, as figure~\ref{fig: correctness}~(left) shows. If we do the same for the first convolutional layer, where we saw lower reasons strengths, figure~\ref{fig: correctness}~(right) indeed does not show such clusters.

\subsection{Improving reasons: do good reasons lead to more robustness and fairness?}
\label{ssec: improving reasons}

We used the reasons method to interpret a trained model, but can we also improve the model's reasons? So the model not just performs well, but does so, literally, ``for the right reasons''~\cite[p.~5192]{Bender2020}? The suggestive hope is that this delivers on the needs for interpretability: if the model has good reasons for its output, it should be more robust and fair.

\paragraph{Training reasons}
To improve a model's reasons via backpropagation, we need a loss function to measure the quality of its reasons with its current weights. Given weights $w$ and a batch $x = (x_1, \ldots, x_N)$ of inputs with corresponding labels $y = (y_1, \ldots, y_N)$, we define the \emph{doxastic reasons loss} $L(w, x,y)$. Let $\{ l_1, \ldots, l_C \}$ be the set of classes (for MNIST this is the set of digits). Let $\hat{y}^w_k$ be the $C$-dimensional vector of logits produced by the model on input $x_k$ using its weights $w$. We want that the $d$-th output neuron is a `good' reason for label $l_d$. To formalize that, define $W := \{ (x_k,y_k) : k = 1, \ldots, N\}$ as the set of worlds. For $d = 1, \ldots, C$, the reasons vector of the $d$-th output neuron is $r_d = (\hat{y}^w_k\,_d : k = 1, \ldots, N)$ and $A_d = \{ (x,y) \in W : y = l_d \}$ is the proposition that the input has label $l_d$. The strength with which $r_d$ speaks for $A_d$ should be high, so
\begin{equation}
    \label{eqn: doxastic reasons loss}
    L (w,x,y) = \sum_{d = 1}^C e^{-\dox(r_d, A_d, b, W)}.
\end{equation}
We instantiate a LeNet model and train it using the sum of the usual loss (i.e., cross entropy) and this reasons loss. For comparison, we make a copy of the initial model and train it on the very same sequence of batches but with only the usual loss. Both models achieve $>99\%$ accuracy. While correctness only marginally improves, faithfulness improves more: in the more difficult `neg2pos' version, the model now achieves success rates between 60\% and 80\% for all digits (previously 6 digits were below $60\%$). More details are in appendix~\ref{app: exp 2}.\footnote{\label{ftn: alternative loss function} There, we also consider an alternative loss function, which we call the \emph{elementary reasons loss}. Curiously, it improves faithfulness and correctness more than the doxastic reasons loss, but it does not improve robustness unlike the doxastic reasons loss (as we will see next). So these notions interact with reasons nontrivially.}

\paragraph{Robustness}
We can improve a model's reasons structure via training. But do good reasons make it harder to trick the model? To test this, we adversarially attack both the reasons-trained model and the comparison model with a FGSM attack~\cite{Goodfellow2015}. This adds $\epsilon$-much adversarially crafted noise to the input images. We check, for different choices of $\epsilon$, how much accuracy decreases due to these attacks. We find that the model trained for reasons is, for all considered $\epsilon$'s, more immune to FGSM attacks than the comparison model. For $\epsilon = 0.15$, this is $78.6\%$ vs $69.9\%$, and for $\epsilon = 0.25$, this is $44.6\%$ vs $27.1\%$ (more details in appendix~\ref{app: exp 2}). This is remarkable: First, nothing in the reasons training is specific to defending adversarial attacks.\footnote{The point here also is not to introduce a new defense to adversarial attacks. Rather, we wanted to answer whether improved reasons lead to more robustness.} Second, the reasons method increases interpretability and robustness while maintaining the same high accuracy. Thus, it defies general tradeoffs between accuracy and interpretability~\cite{Dziugaite2020} and between accuracy and stability~\cite{Colbrook2022}.

\paragraph{Fairness}
Moving to a different modality, we consider the task of predicting whether a person's income is above a given threshold based on tabular data about their age, occupation, sex, etc. We use the modernized \emph{Adult} dataset due to~\cite{Ding2021}, here focusing on US census data from Alabama in 2018. We consider two income thresholds: 25k and 50k. We train multi-layer perceptrons (MLPs) for this task. Treating sex as a protected attribute, we measure the MLPs' fairness using standard metrics: disparate impact (DI)~\cite{Feldman2015} and equality of opportunity (EoO)~\cite{Hardt2016}. We add a reasons-based fairness metric.

Given a list of inputs $x = (x_1, \ldots, x_N)$, let $\hat{y}^w = (\hat{y}^w_1, \ldots, \hat{y}^w_N)$ be the corresponding model outputs computed with its weights $w$. So $\hat{y}^w_k$ is the value of the single output neuron; if the sigmoid of it is $>0.5$, the model predicts the income to be above the threshold. Let $W = \{ x_1, \ldots, x_N\}$ be the set of worlds, let $A_+$ be the set of $x \in W$ which the model predicts to have an income above the threshold, and let $A_p$ (resp., $A_u$) be the set of $x \in W$ that belong to the privileged (resp., unprivileged) group. The reasons vector of the single output neuron is $\hat{y}^w$. To be fair, the model's reasons strength for a positive prediction should be the same regardless of conditioning on the privileged group or the unprivileged group. So, with the uniform measure $b$ on $W$, the \emph{reasons difference} is: 
\begin{equation}
    \label{eqn: fairness loss}
    \RD(w,x) := \big(
        \dox(\hat{y}^w, A_+, b(\cdot|A_p), W) 
        - 
        \dox(\hat{y}^w, A_+, b(\cdot|A_u), W)
    \big)^2.
\end{equation}
This not only measures trained models (smaller is better), but also serves as a loss function to train models (an unsupervised one, since no labels are needed). To do so, we again initialize an MLP, make a copy, train the original model with the sum of the usual loss and the $\RD$ loss, and train the comparison model with only the usual loss (details in appendix~\ref{app: exp 2}). We find that, in the 25k version, the model trained for reasons performs equally well as the comparison model, but it improves on RD. So there is a dimension of fairness in addition to DI and EoO that could still be improved. In the 50k version, the two models get perfect RD scores and perform equally well on the other metrics except DI: here the reasons trained model fares better. This again shows that the comparison model was not yet on the Pareto front of fairness.
In sum, the reasons training could improve along fairness dimensions while keeping the same accuracy---again defying general fairness--accuracy tradeoffs~\cite{Dutta2020}.

\subsection{Reasons in LLMs: mechanistic interpretability}
\label{ssec: reasons in LLMs}

A prominent tool for mechanistic interpretability of large language models (LLMs) are \emph{sparse auto-encoders} (SAEs)~\cite{Cunningham2023, Bricken2023}, which were scaled in~\cite{Templeton2024} to also find abstract features represented by the neural network (e.g., sadness or sycophancy). As mentioned, SAEs are expensive to train and difficult to evaluate, so we test if our reasons method---which only needs forward-passes---also can identify such abstract features. For concreteness, we focus on one abstract feature---\emph{sentiment}---since it is well-studied in NLP with established datasets and baselines.\footnote{For an older discussion of `sentiment neurons' using LSTMs, see~\cite{Radford2017, Donnelly2019}.} We analyze the LLM Qwen2.5-0.5B-Instruct. Although small by today's standards, it solves the sentiment classification task (see below) and can easily be run on a laptop.

\begin{figure}
    \centering
    \begin{minipage}{0.49\linewidth}
        \centering
        \includegraphics[width=\linewidth]{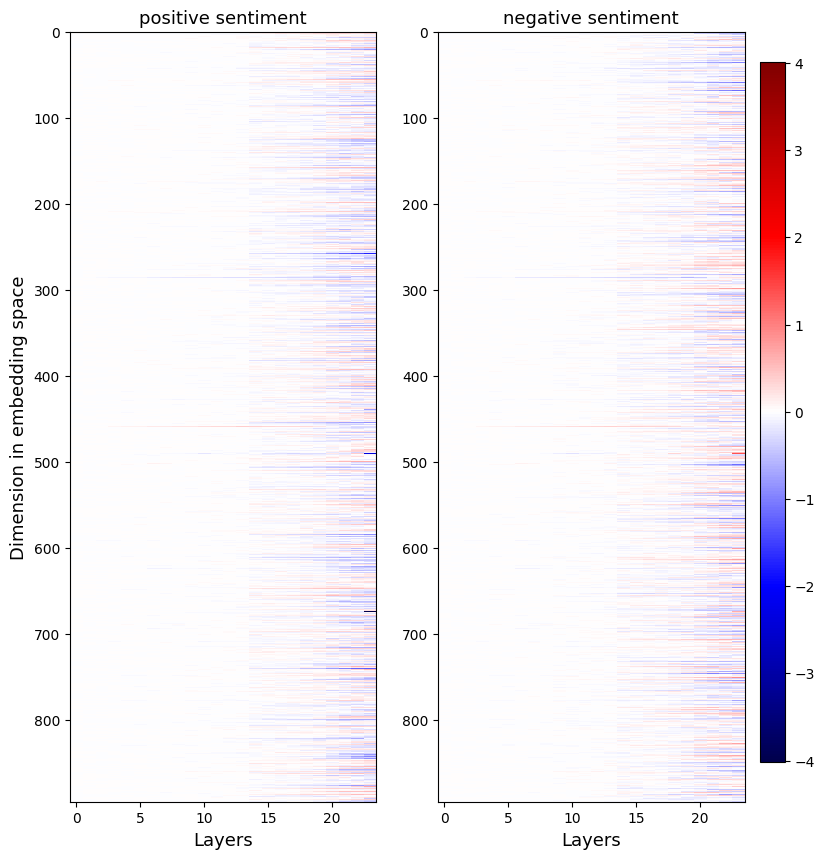}
    \end{minipage}
    \begin{minipage}{0.49\linewidth}
    \scriptsize
    \begin{tcolorbox}[size=small, colframe=teal, colback=lightgray, title=Prompt]
    What do you think of the movie Titanic? Would you recommend watching it? Why or why not?
    \end{tcolorbox}
    \begin{tcolorbox}[size=small, colframe=olive, colback=lightgray, title=Original output]
    The movie Titanic is a classic that has captured the hearts and imaginations of audiences for generations. It tells the story of the RMS Titanic, a luxurious \ldots
    \end{tcolorbox}
    \begin{tcolorbox}[size=small, colframe=red, colback=lightgray, title=Positive intervention]
    Watching Titanic was definitely one of cinema's greatest achievements—it captured audiences worldwide alike through unforgettable visuals and poignant storytelling. \ldots
    \end{tcolorbox}
    \begin{tcolorbox}[size=small, colframe=blue, colback=lightgray, title=Negative intervention]
    Based on varying interpretations and general sentiments based from reviews throughout decades since its inception, perhaps one could reasonably say as much \ldots
    \end{tcolorbox}
    \end{minipage}
    \caption{\emph{Left}: Reason strength of every neuron in the residual stream. \emph{Right}: Generating output with intervention on the `positivity' neurons and the `negativity' neurons, respectively.}
    \label{fig: sentiment in LLMs}
\end{figure}

We identify which neurons in the residual stream of the model speak most strongly for positive and negative sentiment, respectively. To do so, we sample 1024 sentences from the SST2 dataset~\cite{Socher2013}, which contains movie review excerpts (e.g., ``contains no wit, only labored gags''). We use a two-shot prompt template asking about the sentence's sentiment (appendix~\ref{app: exp 3}). The set $W$ of prompts constructed from the selected sentences forms the set of worlds. Now, for each `neuron'---or, rather, position---in the model's residual stream, we can compute its reasons vector: For each position $n$ (of the 896 embedding dimensions) and for each layer $l$ (of the 24 layers), the reasons vector $r_{n,l}$ has, at component $w \in W$, the value $e_n$, where $e$ is the embedding vector for the last token in layer $l$ given prompt $w$. We use the \texttt{NLTK} \emph{SentimentIntensityAnalyzer} to rank the selected sentences by positivity and by negativity. Let $A_+$ (resp., $A_-$) be the set of worlds using the 25 most positive (resp., negative) sentences. Figure~\ref{fig: sentiment in LLMs}~(left) shows, for each position $n$ in layer $l$, the reason strength for positivity $\dox(r_{n,l}, A_+, b, W)$ and for negativity $\dox(r_{n,l}, A_-, b, W)$, respectively (with $b$ the uniform measure on $W$). Most of the strength is again found in the later layers.

Thus, we formed, in an automated way, hypotheses about the roles of the model's neurons. Next, we need to validate if this description of those components is correct~\cite{Sharkey2025}. Here, we again do this via causal interventions (cf.\ section~\ref{ssec: interpreting a classic}). 
We sample 500 sentences from SST2 (different from those used for the worlds). We use a three-shot prompt template to classify a sentence as `a) positive' or `b) negative' (appendix~\ref{app: exp 3}). Thus, the model achieves an accuracy of $91.2\%$. For each sentence, we also perform the following intervention in the last layer (before the unembedding).
\emph{Pos2neg}: if the model classifies the sentence correctly as positive, we set each of the 5 neurons that most strongly speak for positivity to $a' := m - 5a$, where $m$ is the neuron's mean activation and $a$ its current activation; and we set the 5 neurons speaking most against positivity to $a' = m$.  
\emph{Neg2pos}: if the model classifies the sentence correctly as negative, we set the 5 neurons speaking most for negativity to $a' := m - 7a$; and we set the 5 neurons peaking most against negativity to $a' = m$. 
In $97.9\%$ of the cases, the pos2neg intervention indeed flips the model prediction from `positive' to `negative'. The intervention drops the model's average next-token probability for `a' from $67.2\%$ to $27.8\%$. 
In $98.6\%$ of the cases, the neg2pos intervention indeed flips the model prediction from `negative' to `positive'. The average probability for `b' drops from $69.1\%$ to $2.4\%$.

In figure~\ref{fig: sentiment in LLMs}~(right), we can also see this in generation. Given a prompt about the movie \emph{Titanic}, we generate output with the model first as is, then with a positive intervention and then with a negative intervention. The positive (resp., negative) intervention sets the 5 neurons speaking most for positivity (resp., negativity) to $a' = 2m$ (resp., $a' = 20m$) and the 5 neurons speaking most against positivity (resp., negativity) to $a' = m$ (resp., $a' = m$). After the interventions, the output becomes noticeably more positive and negative, respectively, and this can also be observed statistically using the SentimentIntensityAnalyzer across 100 generations (see appendix~\ref{app: exp 3}).

\section{Discussion and conclusion}
\label{sec: discussion and conclusion}

We introduced a new interpretability method based on a formalized notion of reasons. We have shown, both theoretically and empirically, that the method scores well on our desiderata for interpretability. 

\paragraph{Limitations and future work}
As a new method, we established its promise by focusing on breadth rather than depth. Accordingly, future work should continue our experiments in more depth: testing bigger models and harder tasks in experiments~\ref{ssec: interpreting a classic} and~\ref{ssec: reasons in LLMs}, investigating robustness for more adversarial attacks with a comparison to known defenses in experiment~\ref{ssec: improving reasons}, mapping the space of possible loss functions (cf.\ footnote~\ref{ftn: alternative loss function}), and connecting to more metrics and tasks in the algorithmic fairness literature. 
Other questions include: Which theoretical guarantees on faithfulness and correctness can one derive under plausible assumptions? The reasons vectors of the neurons are connected by the model's weights into a \emph{network of reasons}: can one abstract from it a high-level and human-understandable network (analogous to~\cite{Geiger2025}) or identify circuits (analogous to~\cite{Olah2020})?

\bibliographystyle{abbrvnat}
\bibliography{references}

\newpage
\appendix

\section{Experiment 1: Interpreting a classic}
\label{app: exp 1}

\paragraph{Architecture and training details}
The LeNet architecture is as follows:
\begin{itemize}
    \item 
    a convolutional layer of dimension (in-channel: 1, out-channel: 32, kernel: $3\times3$),
    \item
    a convolutional layer of dimension (in-channel: 32, out-channel: 64, kernel: $3\times3$), 
    \item
    a max-pooling layer followed by dropout (25\%) and flattening, 
    \item
    a linear layer to 128 neurons followed by ReLU and dropout (50\%), 
    \item
    a linear layer to the 10 output neurons (for the 10 digits) followed by log-softmax. 
\end{itemize}
We train with a batch size of 64 and 20 epochs, using the AdamW optimizer~\cite{Loshchilov2019} with the default learning rate. We load the MNIST dataset via \texttt{torchvision.datasets}, which contains a balanced 60k images in the training set and 10k images in the test set. 

\begin{figure}
    \centering
    \includegraphics[width=0.6\linewidth]{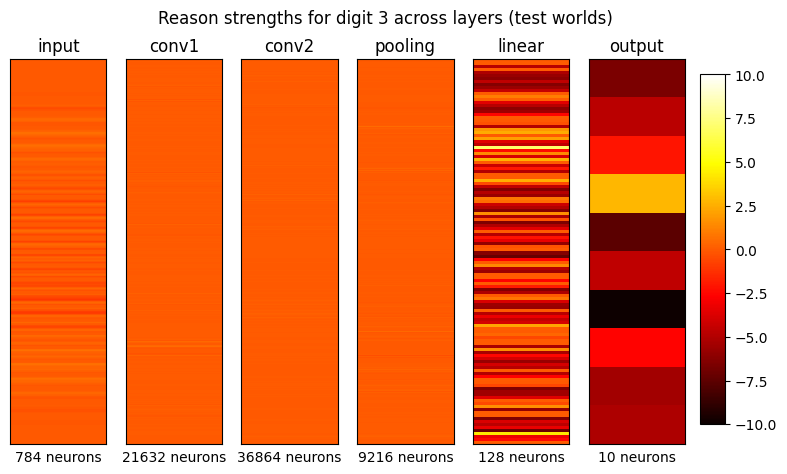}
    \includegraphics[width=0.6\linewidth]{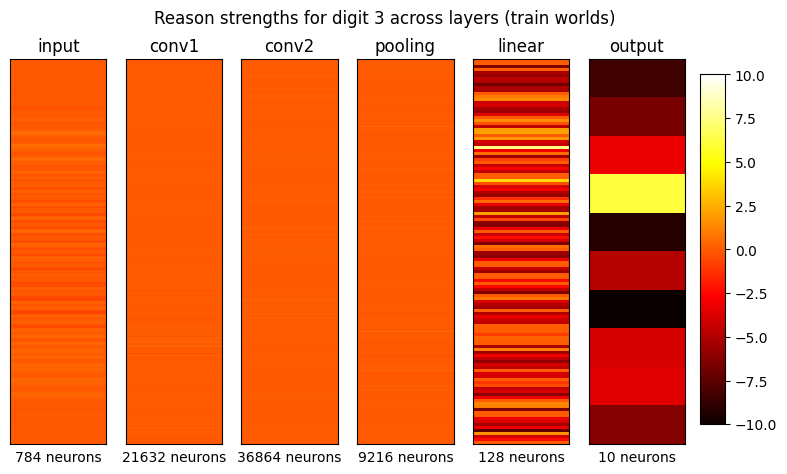}
    \caption{\emph{Top}: Computing reasons strengths using worlds from the test dataset, so the model has not seen them. \emph{Bottom}: Using the training dataset instead.}
    \label{fig: app test vs training worlds}
\end{figure}

\paragraph{Test vs training worlds}
Figure~\ref{fig: app test vs training worlds} shows the difference between using worlds from the test dataset (as done in the main text, so the model has not seen them) and from the training dataset. Using training worlds, the reasons strengths get higher values than with test worlds. After all, the model has seen these worlds during training. However, there is no qualitative difference.

\begin{figure}
    \centering
    \includegraphics[width=0.65\linewidth]{reason_strengths_2025-04-21_04-40-59.png}
    \includegraphics[width=0.65\linewidth]{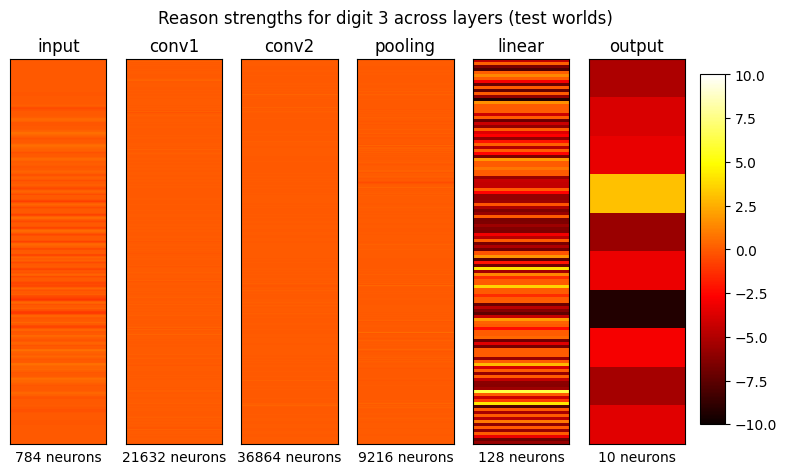}
    \includegraphics[width=0.65\linewidth]{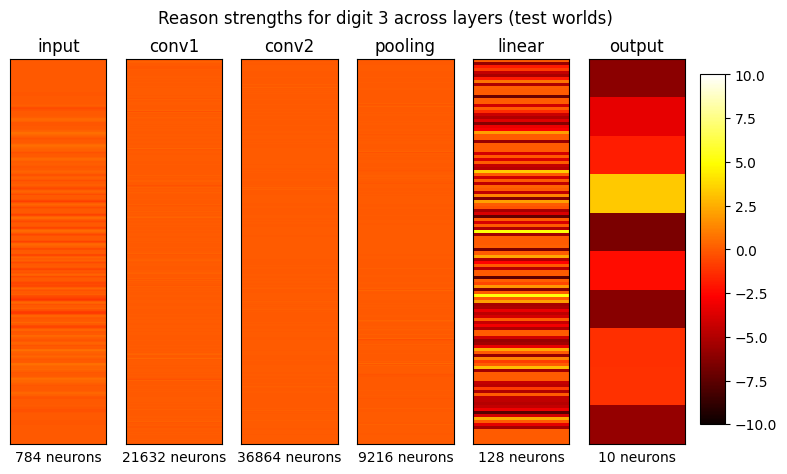}
    \caption{The reason strength of three different models (one plot for one model): all are instances of the LeNet architecture trained on MNIST data but with different seeds.}
    \label{fig: app statistical robustness}
\end{figure}

\paragraph{Stastical robustness}
Figure~\ref{fig: app statistical robustness} shows the statistical robustness of the reasons strengths. We compute the reason strength for three different instances of the LeNet architecture. They all were trained on MNIST data and achieved accuracies  $>99\%$ but used different random seeds. Which neurons in, say, the hidden layer strongly speak for digit 3 and which against can vary between the models. So there is no implicit bias in the LeNet architecture that would force a given neuron in the linear layer to play a specific role with respect to digit classification. However, qualitatively speaking, all models show the same behavior in the output layer: that neuron $d$ strongly speaks for digit $d$ and the others strongly against digit $d$. But, again, the quantity especially of the negative strength can vary.

\paragraph{Groups of neurons}

\begin{figure}
    \centering
    \includegraphics[width=0.7\linewidth]{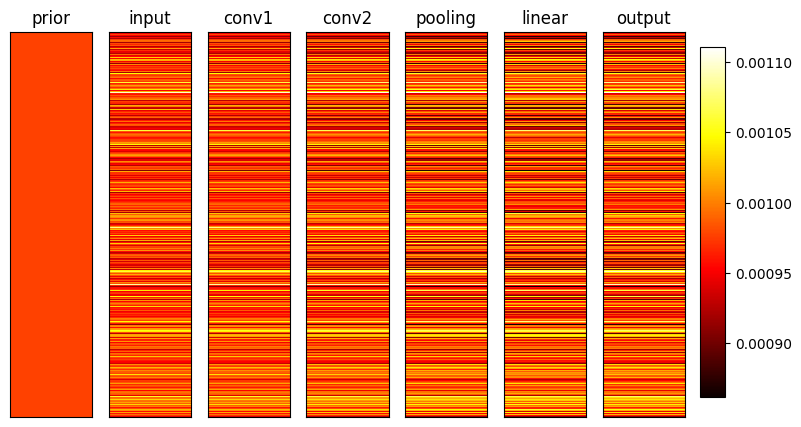}
    \caption{Updating a uniform prior distribution over possible worlds layer by layer.}
    \label{fig: app layerwise update}
\end{figure}

We consider the layers of the trained LeNet model as groups of neurons. We compute how the layers update an initially uniform prior probability distribution over the possible worlds. Specifically, this is done as follows. We again sample 1024 input-label pairs from the MNIST test dataset to form the set of worlds $W$. The prior $b$ is the uniform distribution on $W$. We start with layer~$0$, the input layer. For each neuron in this layer, we compute its reasons vector as before. To aggregate all these reasons vectors, according to the reasons theory, we sum all these vectors, to obtain the reasons vector $r_0$ of layer~$0$. To update the prior probability $b$ with layer~$0$, we compute $b_0 := b * r_0$.\footnote{For numerical stability, we first normalize $r_0$ before computing $b * r_0$.} Similarly, we update $b_0$ with layer~$1$, the first convolutional layer, to obtain $b_2$, and so on for the other layers. The resulting probability distributions are shown in figure~\ref{fig: app layerwise update}. Even though we start with the uniform measure and use a balanced dataset, the input layer introduces some bias among the worlds---and this bias is amplified by later layers.

\begin{figure}
    \centering
    \includegraphics[width=0.49\linewidth]{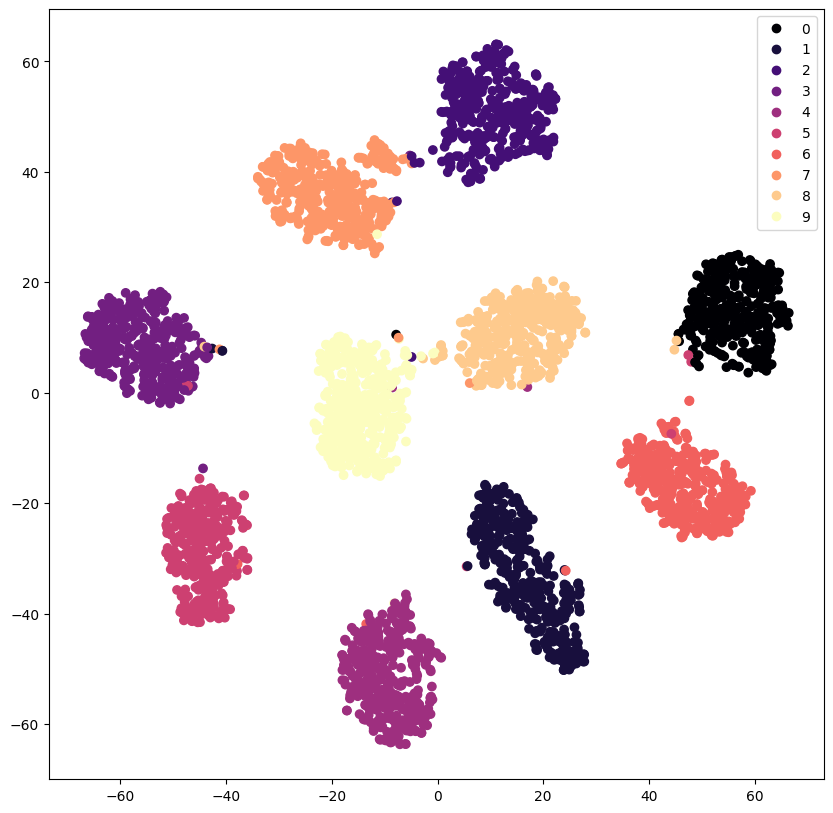}
    \includegraphics[width=0.49\linewidth]{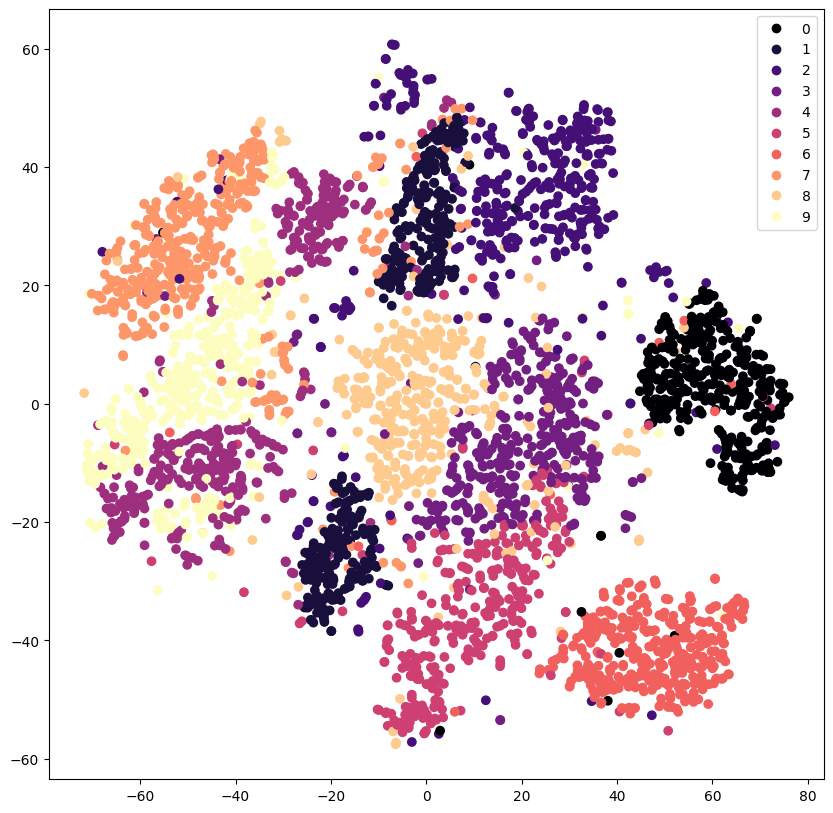}\\
    \includegraphics[width=0.49\linewidth]{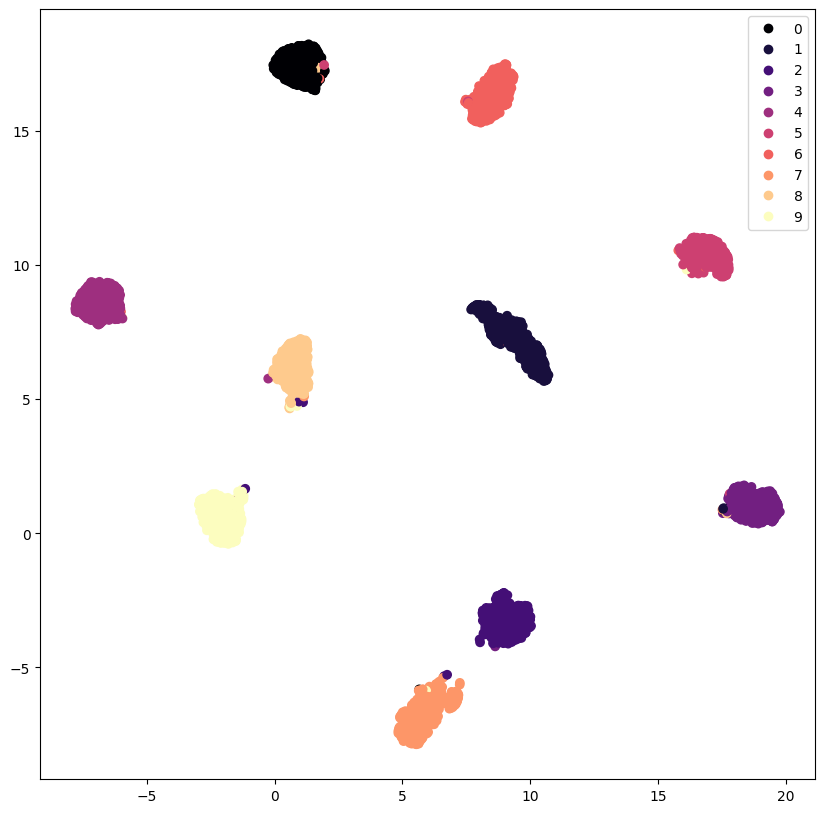}
    \includegraphics[width=0.49\linewidth]{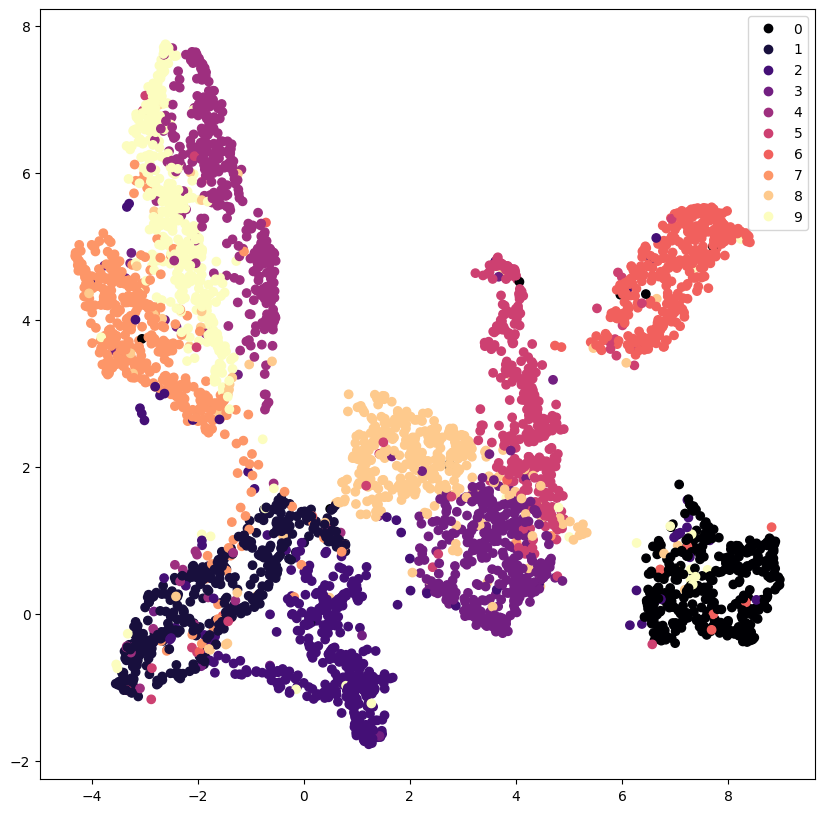}
    \caption{\emph{Top left}: TSNE for linear layer. \emph{Top right}: TSNE for first convolutional layer. \emph{Bottom left}: UMAP for linear layer. \emph{Bottom right}: UMAP for first convolutional layer.}
    \label{fig: app correctness}
\end{figure}

\paragraph{Different dimensionality reduction}
In operationalizing correctness, we used PCA as a dimensionality reduction technique. Other popular such techniques are \emph{t-SNE}~\cite{Maaten2008} and \emph{UMAP}~\cite{McInnes2020}. Figure~\ref{fig: app correctness} shows the results of the correctness experiment when using those dimensionality reductions instead. Qualitatively, the results are the same: clear monochromatic clusters form for the linear layer, but things are more chaotic in the early convolutional layer. For the convolutional layer, the more sophisticated dimensionality reduction methods t-SNE and UMAP can identify more monochromatic clusters compared to PCA, but they are still more chaotic compared to the clear clusters that these methods identify for the linear layer.

\section{Experiment 2: Improving reasons}
\label{app: exp 2}

\paragraph{Training reasons}
After initializing a LeNet model and making a copy, we train the original model with the sum of the usual loss (cross entropy) and the doxastic reasons loss (equation~\ref{eqn: doxastic reasons loss})---adding both summands with equal weight---, while we train the comparison model with only the usual loss. We use the same batches for both models, taken from the MNIST train dataset, with a batch size of 2048 and 20 epochs. We again use the AdamW optimizer~\cite{Loshchilov2019}.

\begin{figure}
    \centering
    \includegraphics[width=0.49\linewidth]{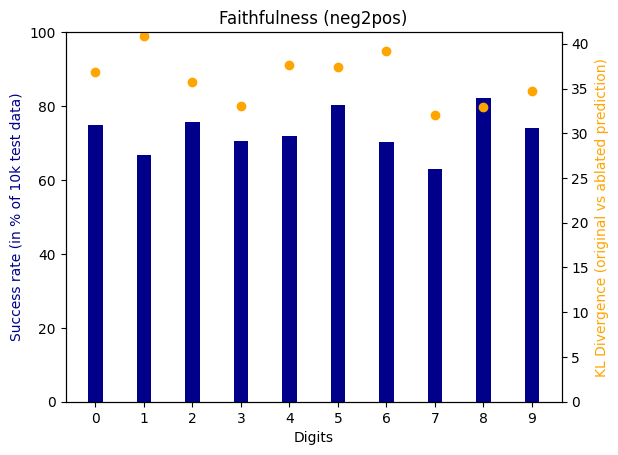}
    \includegraphics[width=0.49\linewidth]{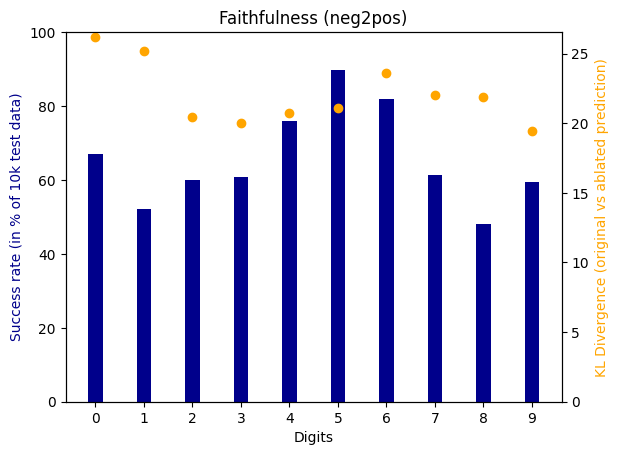}    
    \caption{\emph{Left}: Faithfulness with doxastic reasons training. \emph{Right}: Faithfulness without reasons training.}
    \label{fig: app faithful with and without doxastic reasons training}
\end{figure}

\begin{figure}
    \centering
    \includegraphics[width=0.49\linewidth]{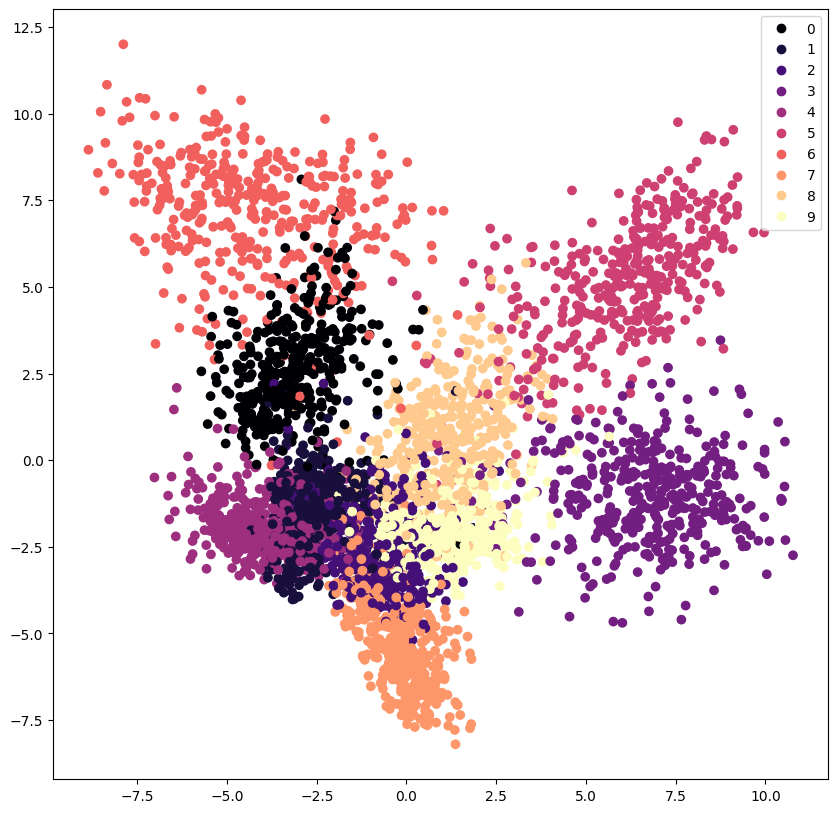}
    \includegraphics[width=0.49\linewidth]{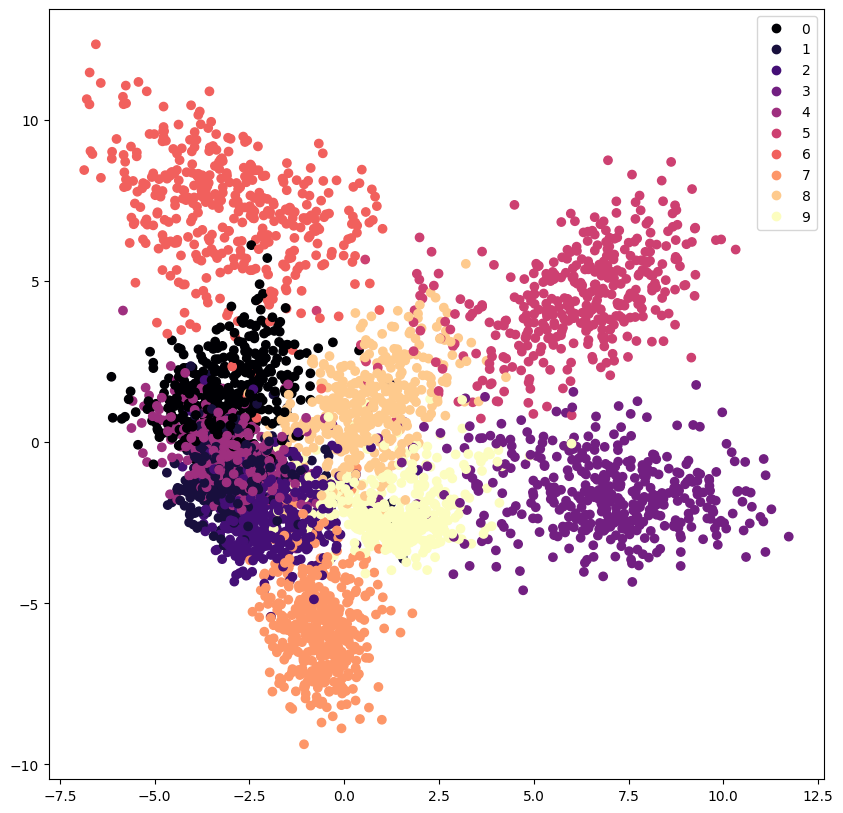}    
    \caption{\emph{Left}: Correctness with doxastic reasons training. \emph{Right}: Correctness without reasons training.}
    \label{fig: app correct with and without doxastic reasons training}
\end{figure}

The achieved accuracy with reasons training is $99.12\%$ on the test set, while without reasons training it is $99.11\%$. Figure~\ref{fig: app faithful with and without doxastic reasons training} shows how the two models compare regarding faithfulness, in the more difficult `neg2pos' version. The success rates now have less variance and are all reliably above $60\%$, and the KL divergences have almost doubled. Figure~\ref{fig: app correct with and without doxastic reasons training} shows how they compare regarding correctness. The clusters only marginally became more separated.

\begin{figure}
    \centering
    \includegraphics[width=0.49\linewidth]{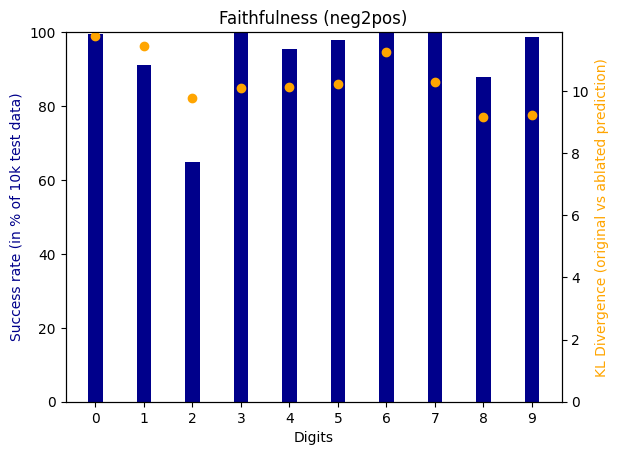}
    \includegraphics[width=0.49\linewidth]{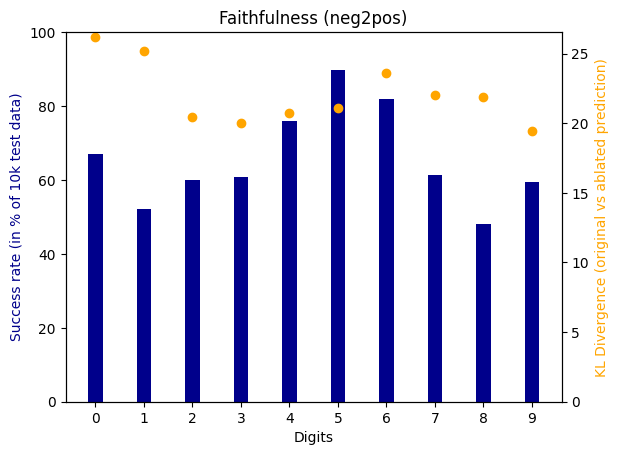}    
    \caption{\emph{Left}: Faithfulness with elementary reasons training. \emph{Right}: Faithfulness without reasons training.}
    \label{fig: app faithful with and without elementary reasons training}
\end{figure}

\begin{figure}
    \centering
    \includegraphics[width=0.49\linewidth]{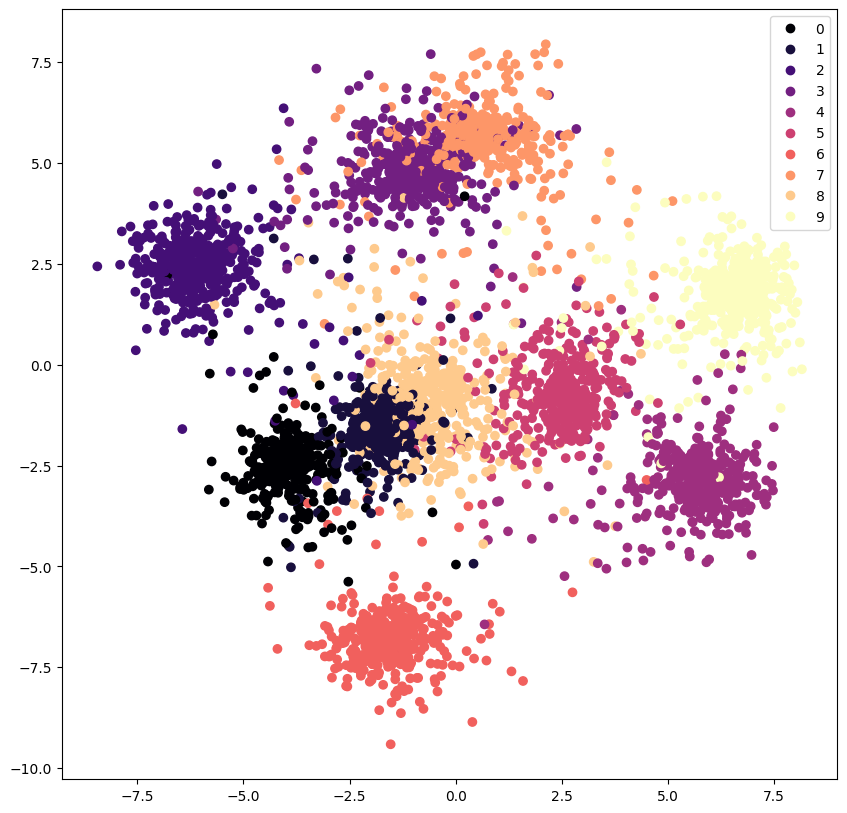}
    \includegraphics[width=0.49\linewidth]{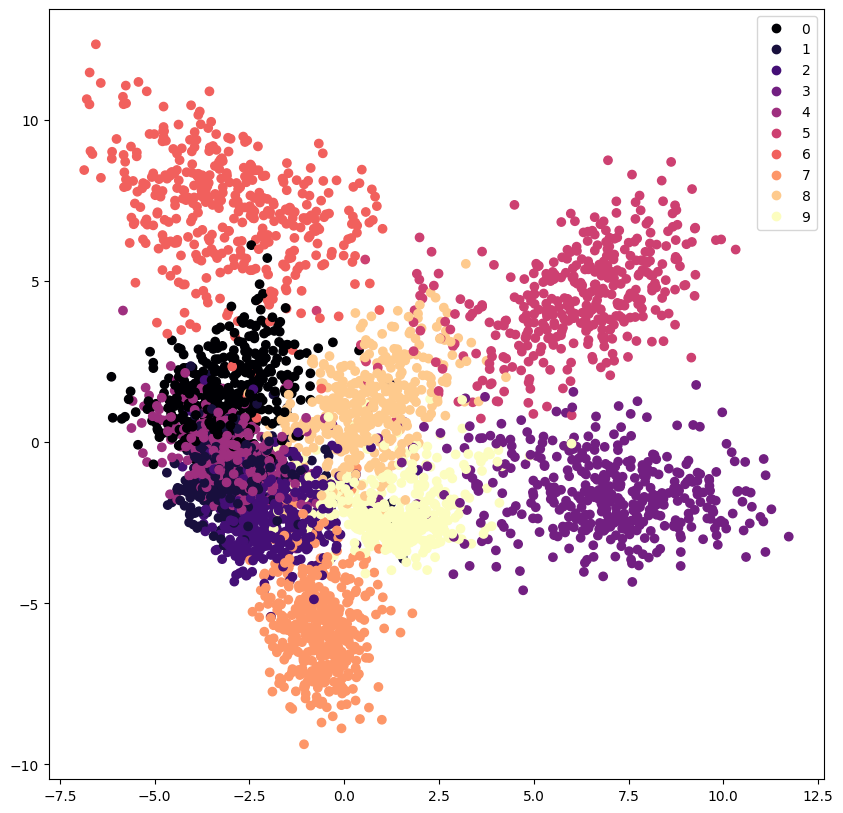}    
    \caption{\emph{Left}: Correctness with elementary reasons training. \emph{Right}: Correctness without reasons training.}
    \label{fig: app correct with and without elementary reasons training}
\end{figure}

\paragraph{Alternative loss function}
Using the same notation as for the doxastic reasons loss defined in equation~\ref{eqn: doxastic reasons loss}, we define here an alternative loss $L'(w,x,y)$, which we call the \emph{elementary reasons loss}. 
Given weights $w$ and a batch $x = (x_1, \ldots, x_N)$ of inputs with corresponding labels $y = (y_1, \ldots, y_N)$, again let $r_d$ be the reasons vector of the $d$-th output neuron, and let $A_d = \{ (x,y) \in W : y = l_d \}$. Another way to formalize that $r_d$ is a `good' reason for $A_d$ is by saying that it is similar to the elementary reason $\mathrm{el}_{A_d}$, which is---in a sense---the canonical reason for $A_d$. We measure similarity by cosine similarity $\mathrm{CosSim}$, which takes values in the interval $[-1, 1]$, where $+1$ means most similar (i.e., codirectional). Hence $1 - \mathrm{CosSim}$ takes values in $[0,2]$ and we want to minimize it. So we define:
\begin{equation}
    \label{eqn: elementary reasons loss}
    L' (w,x,y) := \sum_{d = 1}^C 1 - \mathrm{CosSim}\big(r_d, \mathrm{el}_{A_d} \big).
\end{equation}
When training with this loss, the achieved accuracy with reasons training is $99.05\%$ on the test set, while without reasons training it is $99.11\%$. Figure~\ref{fig: app faithful with and without elementary reasons training} shows how the two models compare regarding faithfulness, in the more difficult `neg2pos' version. The model now achieves, for all digits except 2, a success rate of well above 90\% (compared to around 60\% without reasons training). Figure~\ref{fig: app correct with and without elementary reasons training} shows how they compare regarding correctness. We get yet clearer monochromatic clusters.

\begin{figure}
    \centering
    \includegraphics[width=0.49\linewidth]{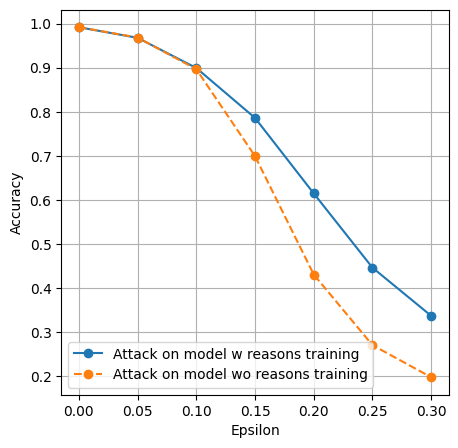}
    \includegraphics[width=0.49\linewidth]{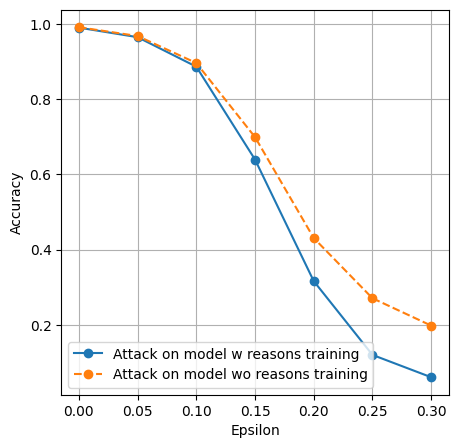}
    \caption{\emph{Left}: Robustness with and without doxastic reasons loss. \emph{Right}: Robustness with and without elementary reasons loss.}
    \label{fig: app robustness doxastic and elementary loss}
\end{figure}

\paragraph{Robustness}
Figure~\ref{fig: app robustness doxastic and elementary loss} shows the effectiveness of an FGSM attack on the reasons trained model compared to the comparison model that has not been trained for reasons. On the left, the reasons training is done with the doxastic reasons loss, and on the right with the elementary reasons loss. Curiously, even though the elementary loss is better at improving the model's reasons structure (as seen in the preceding paragraph) compared to the doxastic loss, it is the doxastic loss which yields more robustness to adversarial attacks, while the elementary loss does not. Thus, there is a nontrivial relationship between faithfulness, correctness, and robustness.

\paragraph{Training fairness via reasons difference}

The task is to predict, based on certain information about a person, whether they earn more than a threshold amount. The two threshold amounts that we test are 25k and 50k. We use the modernized \emph{Adult} dataset due to~\cite{Ding2021}, available via the \texttt{folktables} package.\footnote{Available at \url{https://github.com/socialfoundations/folktables} under the MIT licese.} The data can be chosen to come from different US states and years; here we choose 2018 Alabama. The features in the dataset are the following~\cite[appendix~B]{Ding2021}:
\begin{itemize}
    \item AGEP: Age 
    \item COW: Class of worker
    \item SCHL: Educational attainment
    \item MAR: Marital status
    \item OCCP: Occupation
    \item POBP: Place of birth
    \item RELP: Relationship
    \item WKHP: Usual hours worked per week past 12 months
    \item SEX: Sex (1: Male, 2: Female) 
    \item RAC1P: Recoded detailed race code
\end{itemize}
In this experiment, we will treat sex as the protected attribute. The dataset has 22,268 entries. The percentage of the privileged group (male) is $52.2\%$. In the 25k version, the percentage of positive classification is $62.6\%$; and in the 50k version, it is $31.1\%$.

\begin{figure}
    \centering
    \includegraphics[width=0.7\linewidth]{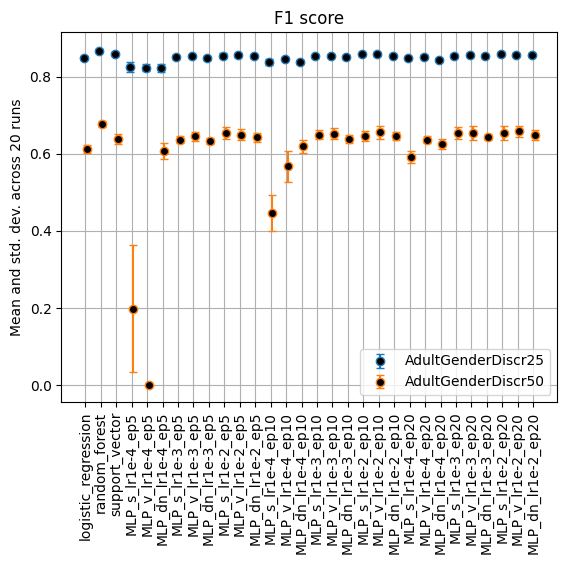}
    \caption{F1 scores for different models for the two fairness tasks.}
    \label{fig: app f1 scores}
\end{figure}

We first train the following models on this task (without any reasons training) to get an understanding of the baseline performance. The first group are the following standard models (available in \texttt{scikit-learn}):
\begin{enumerate}
    \item Logistic regression
    \item Random Forest Classifier
    \item C-Support Vector Classifier
\end{enumerate}
The second group is the following MLPs. Each has 10 input neurons (for the 10 features) and 1 output neuron (indicating positive or negative classification) and uses ReLU as activation function.
\begin{enumerate}
    \item MLP\_s (`small'): One hidden layer of size 100 followed by a second hidden layer of size 50.
    \item MLP\_v (`vanilla'): Four hidden layers each of size 128.
    \item MLP\_dn (`dropnorm'): Also four hidden layers each of size 128, but with $20\%$-dropout and batch norm~\cite{Ioffe2015}.
\end{enumerate}
We test all combinations of the following hyperparameters:
\begin{enumerate}
    \item Learning rates: 1e-4, 1e-3, 1e-2.
    \item Number of epochs: 5, 10, 20. 
\end{enumerate}
We train with binary cross entropy loss (with logits) using AdamW~\cite{Loshchilov2019}. The training-test split is 20\% test data, and we scale the data using \texttt{scikit-learn}'s StandardScaler. Since the dataset is unbalanced, we report the F1 scores (rather than accuracy) achieved by each model in figure~\ref{fig: app f1 scores}.
Except for some outliers on the smallest learning rate, all models achieve a very similar performance.

Based on this, we choose, for further reasons training, the MLP\_dn model with a learning rate of 1e-3 and 20 epochs: it achieves a performance comparable to the other models, but it has higher numerical stability due to batch normalization, which is useful for computing reason strengths (since this requires taking exponentials of neuron activations). 

\begin{figure}
    \centering
    \includegraphics[width=0.7\linewidth]{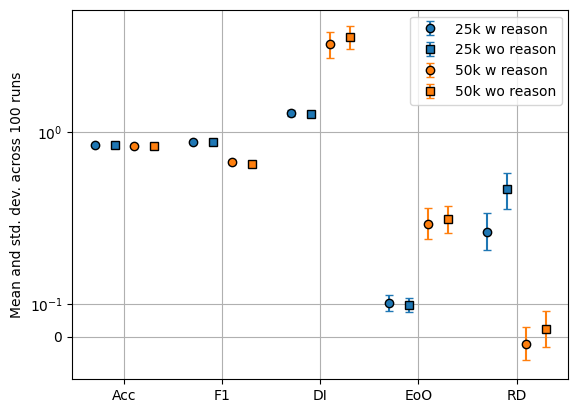}
    \caption{Improving fairness through reasons training. The metrics are: accuracy (Acc), F1 score (F1), disparate impact (DI), equality of opportunity (EoO), reasons difference (RD). Note the logarithmic scale.}
    \label{fig: fairness}
\end{figure}

After initializing the MLP\_dn model and making a copy, we train the original model with the sum of the usual loss (binary cross entropy) and the reasons difference loss (equation~\ref{eqn: fairness loss})---adding both summands with equal weight---, while we train the comparison model with only the usual loss. We use the same batches for both models, with a batch size of 1024, again using AdamW~\cite{Loshchilov2019}.
We do this for both the 25k task and the 50k task, and we repeat each 100 times. 
Figure~\ref{fig: fairness} shows the results. 

\section{Experiment 3: Reasons in LLMs}
\label{app: exp 3}

\paragraph{Setup}
We use the \texttt{nnsight} library~\cite{FiottoKaufman2024} to load the model \emph{Qwen2.5-0.5B-Instruct} and to access its residual stream.\footnote{Available at \url{https://nnsight.net/} under the MIT license.}
We use the popular \emph{Stanford Sentiment Treebank} dataset SST2~\cite{Socher2013}.\footnote{Available at \url{https://huggingface.co/datasets/stanfordnlp/sst2}.}

\paragraph{Prompt-template world construction}
Given a sentence $s$ from the SST2 training dataset, we form the following few-shot prompt:
\begin{itemize}
	\item[]
    \ttfamily
    Input: This was a truly amazing movie.\\
    Classify the sentiment of the message: positive\\
    
    Input: One of the worst films I saw lately.\\
    Classify the sentiment of the message: negative\\
    
    Input: $\{s\}$\\
    Classify the sentiment of the message:\\
\end{itemize}

\paragraph{Prompt-template sentiment classification}
Given a sentence $s$ from the SST2 validation dataset, we form the following few-shot prompt:
\begin{itemize}
	\item[]
    \ttfamily
    You have to classify sentences as either 'positive' or 'negative'.\\
    
    Input: This was a thought-provoking movie \\
    The sentiment of the message is:\\
    a) positive\\
    b) negative\\
    Answer: a)\\
    
    Input: rather mixed acting with a mediocre story line\\
    The sentiment of the message is:\\
    a) positive\\
    b) negative\\
    Answer: b)\\
    
    Input: feel-good story with rich characters\\
    The sentiment of the message is:\\
    a) positive\\
    b) negative\\
    Answer: a)\\
    
    Input: $\{s\}$\\
    The sentiment of the message is:\\
    a) positive\\
    b) negative\\
    Answer:\\
\end{itemize}

\paragraph{Sentiment statistics for generation}

\begin{figure}
    \centering
    \includegraphics[width=0.7\linewidth]{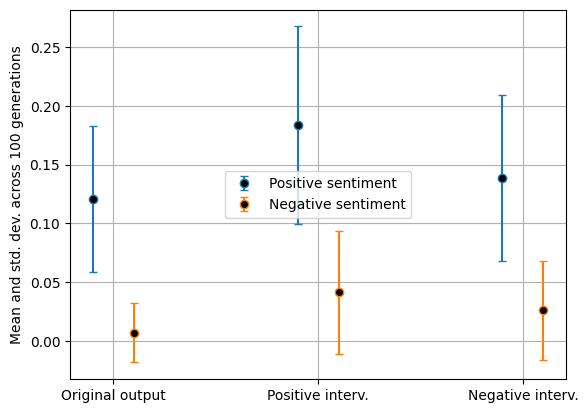}
    \caption{Sentiment statistics when generating 100 responses to the prompt `What do you think of the movie Titanic? Would you recommend watching it? Why or why not?'.}
    \label{fig: app sentiment statistics}
\end{figure}

We consider the prompt `What do you think of the movie Titanic? Would you recommend watching it? Why or why not?'. We generate output with the model first as is (`Original output'), then with a positive intervention (`Positive interv.') and then with a negative intervention (`Negative interv.'). The positive (resp., negative) intervention sets the 5 neurons speaking most for positivity (resp., negativity) to $a' = 2m$ (resp., $a' = 20m$) and the 5 neurons speaking most against positivity (resp., negativity) to $a' = m$ (resp., $a' = m$). To test for statistical variance, we generate these outputs 100 times. For each output, we measure, using the \texttt{NLTK} SentimentIntensityAnalyzer, both its positivity score and its negativity score.\footnote{Available at \url{https://www.nltk.org/index.html} under the Apache-2.0 license.} Figure~\ref{fig: app sentiment statistics} shows the mean and standard deviation of these scores. In figure~\ref{fig: sentiment in LLMs}~(right) in the main text, we saw \emph{qualitatively} that the generated output changes according to the positive or negative intervention. But now we can also see \emph{quantitatively} that the positive intervention generates outputs with higher positivity scores than the original model, and the negative intervention generates outputs with higher negativity scores than the original model.

\section{Compute}
\label{sec: app compute}

All experiments are performed using a regular laptop (CPU with 16~GB memory). The majority of experiments take less than 20~minutes, with a few experiments taking a couple of hours. Initial experiments included approximately 10 days of compute time on the aforementioned setup.

\end{document}